\documentclass{bmvc2k}
\usepackage{graphicx}
\usepackage{amssymb}
\usepackage{booktabs}
\usepackage{bm} 
\usepackage{multirow}
\usepackage{subcaption}
\usepackage[table]{xcolor}
\usepackage{subcaption}

\usepackage[normalem]{ulem}
\useunder{\uline}{\ul}{}
\newcommand{\norm}[1]{\left\lVert#1\right\rVert}

\usepackage{algorithm2e}
\RestyleAlgo{ruled}
\LinesNumbered
\SetKwInput{KwData}{Require}
\newcommand{\imp}[1]{\textcolor{black}{#1}}

\title{Answering from Sure to Uncertain: Uncertainty-Aware Curriculum Learning for Video Question Answering}

\addauthor{Haopeng Li}{hoplee95@outlook.com}{1}
\addauthor{Mohammed Bennamoun}{mohammed.bennamoun@uwa.edu.au}{2}
\addauthor{Jun Liu}{j.liu81@lancaster.ac.uk}{3}
\addauthor{Hossein Rahmani}{h.rahmani@lancaster.ac.uk}{3}
\addauthor{Qiuhong Ke}{qiuhong.ke@monash.edu}{4}

\addinstitution{
 School of Computing and Information Systems,
 The University of Melbourne
}
\addinstitution{
 Computer Science \& Software Engineering, \\
 University of Western Australia
}
\addinstitution{
 School of Computing and Communications, Lancaster University
}

\addinstitution{
 Department of Data Science \& AI,\\
 Monash University
}

\runninghead{Li, Bennamoun, Liu, Rahmani, Ke}{Answering from Sure to Uncertain}

\begin{document}

\maketitle

\begin{abstract}
While significant advancements have been made in video question answering (VideoQA), the potential benefits of enhancing model generalization through tailored difficulty scheduling have been largely overlooked in existing research. This paper seeks to bridge that gap by incorporating VideoQA into a curriculum learning (CL) framework that progressively trains models from simpler to more complex data. Recognizing that conventional self-paced CL methods rely on training loss for difficulty measurement, which might not accurately reflect the intricacies of video-question pairs, we introduce the concept of uncertainty-aware CL. Here, uncertainty serves as the guiding principle for dynamically adjusting the difficulty. Furthermore, we address the challenge posed by uncertainty by presenting a probabilistic modeling approach for VideoQA. Specifically, we conceptualize VideoQA as a stochastic computation graph, where the hidden representations are treated as stochastic variables. This yields two distinct types of uncertainty: one related to the inherent uncertainty in the data and another pertaining to the model's confidence. In practice, we seamlessly integrate the VideoQA model into our framework and conduct comprehensive experiments. The findings affirm that our approach not only achieves enhanced performance but also effectively quantifies uncertainty.
\end{abstract}

%-------------------------------------------------------------------------
\section{Introduction}

Video question answering (VideoQA) has garnered increasing attention from researchers in recent years \cite{huang2020location,seo2021attend,li2022invariant,fan2019heterogeneous,gao2018motion,jiang2020reasoning,9805683,10318951,9576650,9488296,10278436,liu2021question}. Significant efforts have been dedicated to enhancing various aspects of this task, including video encoding \cite{xiao2022video,fan2019heterogeneous}, interaction between video and questions \cite{kim2018bilinear,jiang2020reasoning}, and feature fusion \cite{seo2021attend,gao2018motion}.
Nevertheless, existing works train VideoQA models in a random order, overlooking the fact that optimizing a VideoQA model essentially involves a teaching process, during which the model learns to answer questions of varying difficulties. It has been proven that presenting training examples in a meaningful order, as opposed to random order, can enhance the generalization capacity of models across a wide range of tasks \cite{yang2022hybrid,zhou2022close,soviany2022curriculum}. Such strategies refer to curriculum learning (CL) \cite{bengio2009curriculum} wherein the model is gradually exposed to basic knowledge before advancing to more complex concepts, mimicking human learning. In this work, our goal is to incorporate VideoQA into CL, aiming to enhance the performance of the models.

A major challenge in integrating CL with VideoQA is quantifying data difficulty. Many existing self-paced curriculum learning (SPL) approaches utilize the training loss as a measure of difficulty quantification \cite{kumar2010self,jiang2014easy,zhao2015self,gong2018decomposition}. However, this approach has its shortcomings. Firstly, the training loss generally measures the discrepancy between predictions and ground truth but cannot precisely reflect the inherent difficulty of data. For instance, a math problem could be difficult even if it has a simple answer. Secondly, the training loss varies significantly during training and across different tasks, necessitating a meticulous design of the training scheduler for stable optimization and improved performance \cite{wang2021survey,jiang2014easy,li2017self}.
In response to these challenges, we propose an enhancement to CL for VideoQA by incorporating the principle of uncertainty into the dynamic scheduling of difficulty. We term this approach \textbf{uncertainty-aware curriculum learning (UCL) for VideoQA}. Compared to the training loss, uncertainty offers the advantage of being independent of the ground truth and better reflecting the inherent difficulty of the data. Intuitively, a high-uncertainty video-question pair indicates the presence of potential noise or the model's lack of confidence in its prediction, making it more challenging to handle.
% Instead, in this work, we use uncertainty as the difficulty measurement, which is ground-truth-agnostic and easily bounded.

To quantify data uncertainty and alleviate its negative impact, we propose the utilization of probabilistic modeling for VideoQA. Specifically, we treat VideoQA as a stochastic computation graph \cite{schulman2015gradient}, wherein a video and a question serve as inputs, subsequently undergoing encoding into stochastic representations by a visual encoder and a text encoder. The final predictive distribution of the answer is derived through a combination of the video-question interaction module and the answer prediction module, employing variational inference \cite{wainwright2008graphical,jordan1999introduction}. Within the framework of probabilistic modeling, we define two forms of uncertainty: feature uncertainty, which gauges the intrinsic uncertainty in the data, and predictive uncertainty, which quantifies the model's confidence in its predictions. Notably, our approach to probabilistic modeling and uncertainty quantification remains applicable to both classification and regression tasks. Our contributions can be summarized as follows:
% \begin{itemize}
%     \item We develop a new self-paced curriculum learning framework for VideoQA, where the difficulty of data is measured by the uncertainty that reflects the inherent characteristic of data.
%     \item We propose probabilistic modeling for VideoQA by considering VideoQA as a stochastic computation graph to capture the data uncertainty and to mitigate the negative impact of uncertainty.
%     \item We integrate VideoQA into our uncertainty-aware curriculum learning framework and conduct extensive experiments. The results show that our method achieves state-of-the-art performance on several benchmarks.
% \end{itemize}

1) We develop a self-paced CL framework for VideoQA, where the difficulty of data is measured by the uncertainty that reflects the inherent characteristic of data.
2) We propose probabilistic modeling for VideoQA by considering VideoQA as a stochastic computation graph to capture the data uncertainty and mitigate its impact.
3) We integrate VideoQA into our framework and conduct extensive experiments. The results show that our method achieves better performance and valid uncertainty quantification.

\section{Related Work}

\subsection{Video Question Answering}
\label{vqarw}

Video question answering (VideoQA) is the generalization of visual question answering \cite{9770348,9762027,9830073,cao2021knowledge,ma2021multitask,guo2021bilinear,cao2022bilateral,wang2024bridging,zhang2024latent} from image domain to the video domain.
It requires temporal reasoning over a sequence of events in videos, and various techniques are exploited, such as the attention mechanism \cite{jiang2020divide,jang2019video}, graph neural networks \cite{seo2021attend,park2021bridge}, memory networks \cite{gao2018motion,fan2019heterogeneous,yin2019memory}, and hierarchical structures \cite{le2020hierarchical,xiao2022video}. 
% For example, a dual-LSTM-based approach with both spatial and temporal attention is proposed in \cite{jang2017tgif}. 
% For example, MASN \cite{seo2021attend} models each object as a graph node and captures the spatial and temporal dependencies of all objects with graph neural networks. HQGA \cite{xiao2022video} is developed to model the video as a conditional graph hierarchy to align with the multi-granular nature of questions, which achieves great results on MSVD and MSRVTT \cite{xu2017video}. 
Besides the efforts in improving the model structure, many works develop new frameworks or methodologies for this task \cite{li2022invariant,buch2022revisiting}. 
% For example, invariant grounding is exploited in \cite{li2022invariant} for VideoQA, which aims to find the question-critical scenes whose causal relations with answers are invariant. The atemporal probe (ATP) \cite{buch2022revisiting} is presented to degrade the video-language task to image-level understanding, which provides a stronger baseline on the performance of image-level understanding in the video-language setting than random frames. 
Recently, large-scale video-text pretraining has shown great power in promoting the performance of multimodal video understanding \cite{zellers2021merlot,fu2021violet,zeng2022x}. 
% For instance, MERLOT \cite{zellers2021merlot}, VIOLET \cite{fu2021violet}, and All-in-One \cite{zeng2022x} attain the state-of-the-art performance on several VideoQA benchmarks. 
{However, they require large-scale data and computational resources for training. In this work, we follow \cite{seo2021attend,xiao2022video,buch2022revisiting,li2022invariant} and make comparisons only to the methods without pretraining to show the effectiveness of modeling approaches instead of that of more training data.}
Despite the progress made in VideoQA, {existing works do not consider the impact of the order of training samples or the uncertainty in the data. In this work, we propose a new training framework for VideoQA concerning appropriate difficulty scheduling based on uncertainty.}

\subsection{Curriculum Learning}

Curriculum learning (CL) \cite{bengio2009curriculum} emulates the human learning process by starting with easier tasks and gradually progressing to more challenging ones. Two central components of CL are the difficulty measure and the training scheduler.
In the case of self-paced CL (SPL, where difficulty is measured during training), the loss function is often used as the difficulty measure \cite{kumar2010self,jiang2014easy,zhao2015self,gong2018decomposition}. Initially, during training, samples with higher losses are excluded from optimization. As training advances, the threshold is gradually increased to incorporate more complex data into the optimization process. However, relying solely on loss might not accurately represent the inherent difficulty of data, as difficulty is an intrinsic attribute of samples and should be independent of ground truth labels. To overcome this limitation, we propose employing uncertainty as the difficulty measure for SPL.
To the best of our knowledge, \cite{zhou2020uncertainty} is the only work that also uses uncertainty for CL. However, our method is essentially different from it: 1) We derive uncertainty by probabilistic modeling, while it obtains data uncertainty by a pretrained language model (predefined CL); 2) We perform CL by re-weighting the data, while it adopts baby step \cite{bengio2009curriculum} to arrange data; 3) We focus on VideoQA, while it addresses neural machine translation. The pretrained model and the training scheduler based on baby step make \cite{zhou2020uncertainty} more complex to implement than ours.

\subsection{Uncertainty Modeling}

% In Bayesian modeling, there exist two main types of uncertainty: model (epistemic) uncertainty and data (aleatoric) uncertainty \cite{der2009aleatory,kendall2017uncertainties}. Specifically,  model uncertainty accounts for uncertainty in the model parameters and comes from our ignorance about which model generated the data. This type of uncertainty can be reduced by giving more training data. As for data uncertainty, it captures the inherent noise in our observations such as blur in images and videos. This type of uncertainty is an inherent characteristic of data and cannot be alleviated with more collected data. 
A great number of methods exploit data uncertainty and achieve considerable improvements in various tasks \cite{chang2020data,he2019bounding,yang2021uncertainty,shi2019probabilistic}. 
% For example, DeNet \cite{zhou2021embracing} is proposed to resolve query uncertainty and label uncertainty in temporal grounding, where a decoupling module and a de-bias mechanism are designed for the probabilistic language encoding and diverse temporal regression. UGPT \cite{guo2022uncertainty} is presented for complex action recognition, where the attention scores in Transformer are modeled as probabilistic variables to capture the complex and long-term interaction of actions. 
\imp{Besides, uncertainty has also been applied to CL \cite{chen2022dual,zhou2020uncertainty}. For instance, in the case of \cite{zhou2020uncertainty}, they incorporated data and model uncertainty into the CL process for neural machine translation, focusing on pre-computed data uncertainty to facilitate a baby-step-based CL approach. However, this approach introduces complexity due to the need for prior data uncertainty computation. In contrast, our method stands out by dynamically learning uncertainty during training, which is then utilized to adjust sample weights, leading to a more straightforward and practical implementation. Moreover, the work presented in \cite{chen2022dual} utilized snippet-level uncertainty to assign varying weights to different snippets in the context of weakly-supervised temporal action localization. This was accomplished through the lens of evidential deep learning \cite{amini2020deep,sensoy2018evidential}, with a focus on addressing intra-action variation. Our approach, however, distinguishes itself by introducing probabilistic modeling for uncertainty quantification, with the primary objective of enhancing the generalization capacity of VideoQA models. More importantly, our framework is designed to be a versatile tool applicable to diverse VideoQA models and to improve their performance.}

\section{The Proposed Method}

\subsection{Uncertainty-Based Curriculum Learning}

\subsubsection{Self-Paced Curriculum Learning Revisit}

Curriculum learning (CL) is a training strategy where the networks are trained from easy data to hard data \cite{kumar2010self,jiang2014easy,zhao2015self,gong2018decomposition}. It mimics the organization of curricula for humans, i.e., starting from basic knowledge to complex concepts. Initially, we introduce the concept of self-paced curriculum learning (SPL) as follows. Given the training set $\left\{(\bm{x}_i,y_i)\right\}_{i=1}^{D}$, where $\bm{x}_i$ is the observation, $y_i$ is the ground truth, and $D$ is the number of training data, SPL aims to minimize the following loss in epoch $e$,
$\mathcal{L}(	\bm{\xi},\bm{w};e)=\frac{1}{D}\sum_{i=1}^D w_il\left(f_{\bm{\xi}}(\bm{x}_i),y_i\right)+R\left(\bm{w};e\right)
\label{ospl}$,
where $f_{	\bm{\xi}}$ is a network parameterized by $\bm{\xi}$, $l(\cdot,\cdot)$ is a loss function (such as MSE and cross-entropy), $\bm{w}=\left\{w_i\right\}_{i=1}^{D}$ is the set of weights ranging from 0 to 1 for training data at epoch $e$, and $R(\bm{w};e)$ is the regularization preventing $w_i$ from dropping to 0. In the original SPL \cite{kumar2010self}, $l_1$ norm is exploited as the regularization, i.e.,
$R\left(\bm{w};e\right)=-\lambda(e)\sum_{i=1}^D w_i$,
where $\lambda(e)$ is a scheduler (increasing function) determining the difficulty changing during training. The optimal $\bm{w}$ can be analytically solved and given as follows,
\begin{equation}
w_i^*=\left\{\begin{matrix}1,&\mathrm{if}\ l\left(f_{\bm{\xi}}(\bm{x}_i),y_i\right)<\lambda(e),\\0,&\mathrm{otherwise.}\\\end{matrix}\right.
\end{equation}
An intuitive explanation of the original SPL is that, in epoch $e$, only the training data whose loss are less than $\lambda(e)$ are used for optimization, where the loss can be regarded as a difficulty measure and $\lambda(e)$ is the threshold. As the training proceeds, $\lambda(e)$ is gradually increased to include more difficult data. 

Recognizing that hard/binary weighting of samples might restrict the flexibility of SPL, soft regularizers have been developed \cite{jiang2014easy,gong2018decomposition}. Nonetheless, these SPL methods still rely on the training loss to determine difficulty \cite{kumar2010self,jiang2014easy,zhao2015self,gong2018decomposition}, which has limitations in two aspects: 1) The training loss typically measures the distance between predictions and ground truth, failing to precisely reflect the inherent data difficulty that should remain independent of ground truth; 2) The training loss varies significantly throughout training and across different tasks, necessitating a more meticulous design of $\lambda(e)$ for better performance.

\subsubsection{Uncertainty-Based Curriculum Learning}
\label{ucl}

To tackle the limitations of conventional SPL approaches, we introduce an approach called uncertainty-based curriculum learning. In this method, we leverage uncertainty to quantify the level of difficulty within the curriculum learning framework. In contrast to relying on the training loss, the utilization of uncertainty offers a measure that remains independent of the ground truth and effectively captures the intricacies of the data. Intuitively, a sample with high uncertainty signifies the potential presence of noise or indicates that the model lacks confidence in its prediction, thereby classifying the sample as more challenging to handle.

Concretely, in our SPL, the loss function in epoch $e$ is defined as follows,
$\mathcal{L}(\bm{\xi};e)=\frac{1}{D}\sum_{i=1}^D w_il\left(f_{\bm{\xi}}(\bm{x}_i),y_i\right)$,
where $w_i$ is computed as $w_i=1-\sigma \left(\frac{U_i}{\lambda (e)}\right)$, 
% \begin{equation}
% w_i=1-\sigma \left(\frac{U_i}{\lambda (e)}\right),
% \end{equation}
and $U_i$ represents the (normalized) uncertainty of the sample $\bm{x}_i$, which will be explained in more detail later. $\sigma(\cdot)$ denotes the Sigmoid function, and $\lambda (e)$ is a monotonically increasing function. At the outset of training, samples with low uncertainty (easier data) are assigned higher weights compared to samples with high uncertainty (more challenging examples). As training progresses (with increasing $e$), the weights of low-uncertainty samples and high-uncertainty samples converge towards equality, ultimately resulting in the involvement of all data.
% This approach effectively fulfills the goal of curriculum learning, enabling training to transition from simpler to more complex instances.

Essentially, the uncertainty $U_i$ (and the weight $w_i$ based on it) proposed in this work is a function of network parameters. Consequently, it would also possess gradients with respect to these parameters during the backward propagation process. However, this should be prevented because the model would predict the results that have no uncertainty eventually if no constraint is applied to the weight $w_i$. Previous SPL methods prevent this by applying a regularization to the weight as shown in the previous section. Nevertheless, we lack knowledge of the uncertainty prior across the training set, making it challenging to determine an appropriate weight regularization. To overcome this hurdle, we take an alternative approach: we detach the weight from the computation graph and nullify the gradients with respect to the parameters, i.e.,
% \begin{equation}
% \frac{\partial w_i}{\partial \xi}:=0, \forall i=1,\cdots,D.
% \end{equation}
$\nabla_{\bm{\xi}} w_i:=\bm{0}, \forall i=1,\cdots,D$.
% The reason why we set the gradient to 0 instead of applying a regularization as in Eq. \ref{ospl} is that we do not know the prior of uncertainty on the whole training set, and so it is intractable to find an appropriate regularization to the weight.

\subsection{Probabilistic Modeling for VideoQA}
\label{pm}

% \begin{figure}[tbp]
% \centering
% \includegraphics[width=\columnwidth]{scg.pdf}
% \caption{The stochastic computation graph for VideoQA. The hidden representations of the input video $V$ and question $Q$ are modeled as stochastic nodes ($M$ and $N$). The predictive distribution $Y$ is derived by variational inference over the graph.}
% \label{scg}
% \end{figure}

A remaining challenge for our UCL is the uncertainty quantification. To tackle this challenge, we propose probabilistic modeling for VideoQA. 
Specifically,  we consider VideoQA as a stochastic computation graph, where the input nodes are the video $V$ and the question $Q$. Following video encoding (parameterized by $\phi$) and question encoding (parameterized by $\psi$), we obtain the  stochastic video representation $M$ and the  stochastic question representation $N$. 
The distribution of the answer $Y$ can be deduced by variational inference:
\begin{equation*}
       p_{\phi,\psi,\theta}(y|V,Q) =\int_{m,n}p_{\phi,\psi,\theta}(y,m,n|V,Q)\mathrm{d}m\mathrm{d}n \notag
    %   &=\int_{m,n}\frac{p(y,m,n,V,Q)}{p(V,Q)}\\
    %   &=\int_{m,n}\frac{p(y|m,n)p(m|V)p(n|Q)p(V)p(Q)}{p(V)p(Q)}\\
    %   &=\int_{m,n} p_{\theta}(y|m,n)q_{\phi,\psi}(m,n|V,Q)\mathrm{d}m\mathrm{d}n\notag\\
       =\int_{m,n} p_{\theta}(y|m,n)q_{\phi}(m|V)q_{\psi}(n|Q)\mathrm{d}m\mathrm{d}n.
\end{equation*}
Since the integral over $(m,n)$ is intractable, we sample $m$ and $n$ from $q_{\phi}(m|V)$ and $q_{\psi}(n|Q)$, respectively, for $K$ times to approximate the predictive distribution of $Y$, i.e.,
$p_{\phi,\psi,\theta}(y|V,Q)\approx\frac{1}{K}\sum_{k=1}^Kp_{\theta}(y|m_k,n_k),\label{output}$
where $m_k\sim q_{\phi}(m|V)$ and $n_k\sim q_{\psi}(n|Q)$. In practice, we use the reparameterization trick \cite{kingma2013auto} to make the sampling differentiable.
$p_{\theta}(y|m_k,n_k)$ can be specified for classification and regression as follows,
\begin{equation}
\textnormal{\textit{cls}: } p_{\theta}(y|m_k,n_k)=\mathrm{softmax}\left(g_k\right),\;
\textnormal{\textit{reg}: } p_{\theta}(y|m_k,n_k)=\mathcal{N}(\mu_k,\sigma^2_k),
\end{equation}
where $g_k=g(m_k,n_k;\theta)\in\mathbb{R}^C$ is the predicted logits ($C$ is the number of classes), and $\mu_k=\mu(m_k,n_k;\theta)$ and $\sigma_k^2=\sigma^2(m_k,n_k;\theta)$ are the predicted expectation and variance.

% The proposed stochastic computation graph for VideoQA cannot be directly optimized by maximising the log-likelihood of the predictive distribution as traditional deep neural networks because the analytical solution of the integral is intractable. Therefore, we instead maximize the evidence lower bound (ELBO) that is derived as follows,
The proposed stochastic computation graph for VideoQA is optimized by maximizing the evidence lower bound (ELBO) that is derived as follows\footnote{The derivation can be found in the supplementary material.},
\begin{align*}
\log p_{\theta,\phi,\psi}(y|V,Q)
% =&\log \int_{m,n} p_{\theta,\phi,\psi}(y,m,n|V,Q)\mathrm{d}m\mathrm{d}n \\
% =&\log \int_{m,n} p_{\theta,\phi,\psi}(y,{m,n}|x)\frac{q_{\phi}({m,n}|V,Q)}{q_{\phi}({m,n}|V,Q)}\mathrm{d}m\mathrm{d}n\\
% &=\log \mathbb{E}_{(m,n)\sim q_{\phi,\psi}({m,n}|V,Q)}\left[\frac{p_{\theta,\phi,\psi}(y,m,n|V,Q)}{q_{\phi,\psi}(m,n|V,Q)}\right] \\
&\geq \mathbb{E}_{(m,n)\sim q_{\phi,\psi}({m,n}|V,Q)}\left[\log\frac{p_{\theta,\phi,\psi}(y,m,n|V,Q)}{q_{\phi,\psi}(m,n|V,Q)}\right] \\
%  =& \mathbb{E}_{(m,n)\sim q_{\phi,\psi}({m,n}|V,Q)}\left[\log\frac{p_{\theta}(y|m,n)p(m,n)}{q_{\phi,\psi}(m,n|V,Q)p(V,Q)}\right] \\
%  = &\mathbb{E}_{(m,n)\sim q_{\phi,\psi}({m,n}|V,Q)}\left[\log p_{\theta}(y|m,n)-\log\frac{q_{\phi,\psi}(m,n|V,Q)}{q(m,n)}-\log p(V,Q)\right] \\
%  =& \mathbb{E}_{(m,n)\sim q_{\phi,\psi}({m,n}|V,Q)}\left[\log p_{\theta}(y|m,n)\right]-\mathbb{E}_{(m,n)\sim q_{\phi,\psi}(m,n|V,Q)}\left[\log\frac{q_{\phi,\psi}(m,n|V,Q)}{q(m,n)}\right]\\
% &-\mathbb{E}_{z\sim q_{\phi,\psi}(m,n|V,Q)}\left[\log p(V,Q)\right]  \\
% =& \mathbb{E}_{(m,n)\sim q_{\phi,\psi}({m,n}|V,Q)}\left[\log p_{\theta}(y|m,n)\right]-D_{\mathrm{KL}}\left(q_{\phi,\psi}(m,n|V,Q)||p(m,n)\right)-\log p(V,Q)\\
% =& \mathbb{E}_{(m,n)\sim q_{\phi,\psi}({m,n}|V,Q)}\left[\log p_{\theta}(y|m,n)\right]-D_{\mathrm{KL}}\left(q_{\phi}(m|V)q_{\psi}(n|Q)||p(m)p(n)\right)-\log p(V)p(Q)\\
&= \mathbb{E}_{(m,n)\sim q_{\phi,\psi}({m,n}|V,Q)}\left[\log p_{\theta}(y|m,n)\right]-\log p(V) p(Q)\\
&-D_{\mathrm{KL}}\left(q_{\phi}(m|V)||p(m)\right)-D_{\mathrm{KL}}\left(q_{\psi}(n|Q)||p(n)\right),
\end{align*}
where $D_{\mathrm{KL}}\left(\cdot||\cdot\right)$ is the Kullback–Leibler (KL) divergence. 
% The KL divergence can also be regarded as regularization to $M$ and $N$, preventing them from degradating to deterministic representations.
Since $p(V)$ and $p(Q)$ are irrelevant to optimization, once we apply normal Gaussian prior to $M$ and $N$ ($q_{\phi}(m|V)=\mathcal{N}(\mu_m,\sigma^2_m),q_{\psi}(n|Q)=\mathcal{N}(\mu_n,\sigma^2_n)$), the objective can be re-written and approximated as
\begin{align*}
\mathcal{L(\theta,\phi,\psi)}=&-\frac{1}{K}\sum_{k=1}^K \log p_{\theta}(y|m_k,n_k)+\frac{\alpha}{2}\sum_d (1+\log (\sigma_m)_d^2-(\mu_m)_d^2-(\sigma_m)_d^2)\\
&+\frac{\alpha}{2}\sum_d (1+\log (\sigma_n)_d^2-(\mu_n)_d^2-(\sigma_n)_d^2),
\end{align*}
where $\alpha$ is a hyper-parameter. For classification/regression, $\log p_{\theta}(y|m_k,n_k)$ is specified as 
\begin{equation*}
\textnormal{\textit{cls}: } \log p_{\theta}(y|m_k,n_k)=\log p^k_{c^*},\;\textnormal{\textit{reg}: } \log p_{\theta}(y|m_k,n_k)=-\frac{1}{\sigma_k^2}(\mu_k-y)^2-\log\sigma_k^2-\log 2\pi,
\end{equation*}
where $p^k_{c^*}$ is the predictive probability for the correct class $c^*$, and $y$ is the ground truth value.

We then define two types of uncertainty for difficulty quantification based on our probabilistic modeling: feature uncertainty and predictive uncertainty. Feature uncertainty measures inherent uncertainty in data such as noise, blur and occlusion in videos. This type of uncertainty is computed based on the variance of $p_{\theta}(y|m_k,n_k)$ across different sampling results. Concretely, the feature uncertainty for classification is defined as the variance of predicted logits, while the feature uncertainty for regression is computed as the variance of predicted expectations, i.e., 
% \begin{align}
% &\textnormal{\textit{cls}: } U_F=\frac{1}{CK}\norm{\sum_{k=1}^K(g_k-\bar{g})^2}_1,\label{cls}\\
% &\textnormal{\textit{reg}: } U_F=\frac{1}{K}\sum_{k=1}^K(\mu_k-\bar{\mu})^2,
% \end{align}
\begin{equation}
\textnormal{\textit{cls}: } U_F=\frac{1}{CK}\norm{\sum_{k=1}^K(g_k-\bar{g})^2}_1,\;
\textnormal{\textit{reg}: } U_F=\frac{1}{K}\sum_{k=1}^K(\mu_k-\bar{\mu})^2,\label{cls}
\end{equation}
where $\bar{g}=\frac{1}{K}\sum_{k=1}^Kg_k\in\mathbb{R}^C$ and $\bar{\mu}=\frac{1}{K}\sum_{k=1}^K\mu_k\in\mathbb{R}$ are the average of predicted logits and that of predicted expectations, respectively, and the power in Eq. \ref{cls} is performed element-wisely. Feature uncertainty essentially measures the difference among the predictions from different sampled features. That is to say, if the variances of feature $m$ and $n$ are zeros (i.e., there is no uncertainty in them), the computed feature uncertainty would be zero because the random sampling degrades to a deterministic process. As for the predictive uncertainty, it measures the confidence of the model in the final outputs. Concretely, this type of uncertainty for classification is defined as the entropy of the output distribution, while for regression, we use the predicted variance as the predictive uncertainty, i.e.,
\begin{equation}
\textnormal{\textit{cls}: } U_P=-\sum_{c=1}^Cp_c\log p_c,\;
\textnormal{\textit{reg}: } U_P=\frac{1}{K}\sum_{k=1}^K\sigma^2_k,
\end{equation}
where $p_c=\frac{1}{K}\sum_{k=1}^Kp^k_c$ is the predicted probability of class $c$. 
% The predictive uncertainty is computed based on the final outputs of the model, which is agnostic to the probabilistic modeling. 
A lower predictive uncertainty means the model has more confidence in its prediction. The proposed feature uncertainty and predictive uncertainty depict the inherent characteristics of data and are agnostic to ground truth, which are more reasonable for the difficulty measurement in SPL. 

% \imp{Distinguishing itself from aleatoric and epistemic uncertainty, our uncertainty serves a distinct purpose: our feature uncertainty assesses the uncertainty linked to high-level video features, while predictive uncertainty conveys the model's confidence in its predictions. In contrast, aleatoric uncertainty quantifies uncertainty within observations, while epistemic uncertainty pertains to uncertainty in model parameters. {Importantly, it's worth noting that aleatoric and epistemic uncertainty can also be incorporated into our UCL framework. We substantiate this further by presenting a comparison of various uncertainties in the supplementary.}}

Despite that the above two types of uncertainty can measure the difficulty of data, they cannot be directly applied to SPL because their ranges are intractable, which makes it difficult to determine suitable schedulers for SPL (similar to the design of $\lambda(e)$ in the original SPL). To address this issue, we assume they are Gaussian distributed and normalize them to the standard Gaussian distribution, i.e., $\bar{U}=\frac{U-\mathrm{E}[U]}{\sqrt{\mathrm{Var}[U]}}$,
where $U$ represents $U_F$ or $U_P$, and $\mathrm{E}[U]$ and $\mathrm{Var}[U]$ are the mean and the variance of the uncertainty $U$, respectively. Consequently, $\bar{U}_F,\bar{U}_P\sim \mathcal{N}(0,1)$ and they are statistically bounded. Since the mean and variance for the whole dataset are intractable, we use batch normalization \cite{ioffe2015batch}  as a practical means to estimate the normalized uncertainty. In this work, we exploit the two types of normalized uncertainty for SPL. To maintain simplicity, we still use $U$ to denote normalized $U_F$ or $U_P$\footnote{The pseudo code is provided in the supplementary material}.

\subsection{Uncertainty-Aware Curriculum Learning for VQA}

% The proposed framework is agnostic to the essential modules in VideoQA models (i.e., video encoder, question encoder, video-question interaction module, and answer decoder). Therefore, it can be applied to existing VideoQA methods as a plug-and-play method. 
In this section, we show how a VideoQA model can be adapted into our framework. Specifically, we choose MASN \cite{seo2021attend} for our purpose because 1) it is a typical VideoQA model with clear VideoQA modules; 2) it uses various types of visual features that may contain more uncertainty in the hidden space. 

% We first revisit MASN from its input and model structure. MASN exploits four types of visual features, i.e., the global/local appearance/motion feature. Specifically, the pretrained Faster-RCNN \cite{ren2015faster} is used to obtain the object bounding boxes and their RoI features in every frame. The RoI features are regarded as the local appearance features. The global appearance feature is computed as the spatial average of the activation map from the last convolution layer in the object detector. As for motion features, for each frame, its adjacent eight frames (including itself) are input into the pretrained I3D \cite{carreira2017quo}. The extracted features are 2D feature maps (the temporal dimension is 1 when the input is an 8-frame clip). The global motion feature of each frame is computed as the spatial average of its motion feature map, while the local motion feature is obtained by applying RoIAlign \cite{he2017mask} to the feature map with the bounding boxes. 
We first follow the process in MASN to obtain the visual features.
Concretely, for a video of $T$ frames, if $N$ objects are detected in each frame, we can obtain the global appearance features $\left\{\bm{g}_a^{t}\right\}_{t=1}^{T}$, global motion features $\left\{\bm{g}_m^{t}\right\}_{t=1}^{T}$, local appearance features $\left\{\bm{l}_a^{it}\right\}_{i=1,t=1}^{N,T}$, and local motion features $\left\{\bm{l}_m^{it}\right\}_{i=1,t=1}^{N,T}$, where $\bm{g}_a^t\in\mathbb{R}^{1024}$ and $\bm{g}_m^t,\bm{l}_a^{it},\bm{l}_m^{it}\in\mathbb{R}^{2048}$.
% There are two parallel streams in the video encoder of MASN for appearance feature encoding and motion feature encoding, respectively. Since the two streams are identical in structure, we elaborate on them without specifying appearance or motion. 
The visual encoders in MASN forappearance feature and motion feature are essentially two graph convolution networks (GCNs) \cite{welling2016semi} where the object features are modeled as a graph. Specifically, the spatial information (coordinates of bounding boxes) and the temporal information (the encoding of frame indexes) are first fused with the local features. Similarly, the temporal information is fused with the global features. The final node representations are the local features combined with global features. Linear projections are applied to the above feature fusions. Concretely, the node $\bm{n}^{it}$ is computed as follows,
\begin{equation}
\hat{\bm{l}}^{it} = \mathrm{ReLU}\left(\bm{W}_l\left[\bm{l}^{it},\bm{b}^{it},\bm{s}(t)\right]\right),\;
\hat{\bm{g}}^{t} = \bm{W}_g\bm{g}^{t}+\bm{s}(t),\;
\bm{n}^{it} = \mathrm{ReLU}\left(\bm{W}_n\left[\hat{\bm{l}}^{it},\hat{\bm{g}}^{t}\right]\right),
\end{equation}
where $\bm{b}^{it}\in\mathbb{R}^4$ is the coordinate of the bounding box of the $i$-th object in the $t$-th frame, $\bm{s}(t)$ is the sinusoidal function for time-step encoding \cite{vaswani2017attention}, $[\cdot]$ represents  concatenation, and $\bm{W}_l,\bm{W}_g,\bm{W}_n$ are parameters. For simplicity, we flatten the two indexes of $\bm{n}^{it}$ and denote them as $\left\{\bm{n}^{z}\right\}_{z=1}^{Z}$, where $Z=NT$ is the number of all objects in the video.

We combine all the nodes $\left\{\bm{n}^{z}\right\}_{z=1}^{Z}$ to form a object matrix $\bm{N}\in\mathbb{R}^{Z\times d}$, where $d$ is the dimension of the node representations. The adjacent matrix for the graph is computed based on the dot-product of the projected nodes, i.e, $\bm{A}=\mathrm{softmax}\left((\bm{N}\bm{W}_1)(\bm{N}\bm{W}_2)^{\mathrm{T}}\right)$, where $\bm{W}_1,\bm{W}_2$ are parameters. The visual features are then encoded by a GCN, i.e., $\bm{M}=\mathrm{GCN}(\bm{N},\bm{A})$, where $\bm{M}\in\mathbb{R}^{Z\times d}$ contains interaction information among objects. 
% \begin{equation}
% \bm{M}=\mathrm{GCN}(\bm{N},\bm{A}).
% \end{equation}
% The obtained $\bm{M}\in\mathbb{R}^{NT\times d}$ contains interaction information among objects. 

In this work, we encode the video to a probabilistic representation, which is defined as the combination of all encoded objects $\mathcal{R}=\left\{\bm{r}^{z}\right\}_{z=1}^{Z}$. $\bm{r}^{z}$ is the encoded $z$-th object in the video, which is modeled as a multivariate Gaussian distribution, i.e., $\bm{r}^{z}\sim\mathcal{N}(\bm{\mu}^{z},\bm{\sigma}^{z}\bm{I}_d)$, where $\bm{\mu}^{z},\bm{\sigma}^{z}\in\mathbb{R}^{d}$ is the expectation and the variance, and $\bm{I}_d$ is the $d$-order identity matrix (we assume the dimensions in $\bm{r}^{z}$ are independent). The expectation and variance are obtained from $\bm{M}$ with a two-layer perceptron (MLP) as follows, $\bm{\mu}^{z},\bm{\sigma}^{z}=\mathrm{MLP}(\bm{M}^{z})$, where $\bm{M}^{z}\in\mathbb{R}^{d} $ is the $z$-th row in $\bm{M}$. In practice, we use the reparameterization trick to sample $K$ sets of visual representations $\left\{\mathcal{R}_k\right\}_{k=1}^{K}$, where $\mathcal{R}_k=\left\{\bm{r}_k^{z}\right\}_{z=1}^{Z}$ and $\bm{r}_k^z=\bm{\mu}^{z}+\bm{\sigma}^{z}\odot\bm{\epsilon}_k^{z}$ ($\bm{\epsilon}_k^{z}\sim\mathcal{N}(\bm{0},\bm{I}_d)$). $\left\{\mathcal{R}_k\right\}_{k=1}^{K}$ is then exploited for video-question interaction. Note that the process above is parallel for appearance encoding and motion encoding, and thus two types of visual representations are obtained.

The question encoder consists of word embedding and LSTM \cite{hochreiter1997long}. Specifically, the words in the question are converted to 300D vectors by GloVe \cite{pennington2014glove}. Then a linear projection and LSTM are applied to process the word sequence. For a question of $J$ words, the LSTM encodes it to a sequence of hidden states $\left\{\bm{w}^{j}\in\mathbb{R}^{d}\right\}_{j=1}^{J}$
\footnote{We do not apply probabilistic modeling to question as our prior experiments show it brings no improvement.}.

After obtaining the appearance/motion representation and the question encoding, appearance / motion-question interaction is applied, by which the appearance/motion-question interacted representation is obtained. Finally, the above two cross-modal representations are fused to obtain the video-question feature $\bm{s}\in\mathbb{R}^d$ for answer prediction\footnote{Please refer to the supplementary material for the details about the video-question interaction module and appearance-motion fusion module.}. Since we have $K$ sets of appearance/motion representations from sampling, we compute $K$ features $\left\{\bm{s}_k\right\}_{k=1}^{K}$.
% The answer decoder is different for different tasks. 
For the regression task, the answer is predicted by projecting $\bm{s}$ to two scalars (predicted expectation and variance). For the classification task, the answer distribution is computed by projecting $\bm{s}$ to logits and applying Softmax to the logits. For the multi-choice task, we first concatenate the question and each answer, and then model VideoQA as binary classification. The prediction is the answer with the highest correct probability. 
% Note that $\left\{\bm{s}_k\right\}_{k=1}^{K}$ would predict $K$ results, and the final result is computed as their average. 
% The optimization follows the uncertainty-aware curriculum learning proposed in Section \ref{ucl}.

\section{Experiments}
\subsection{Implementation Details}

In the probabilistic modeling, the visual representations are sampled five times during training and ten times during testing. As for the CL scheduler $\lambda(e)$, we employ a linear\footnote{Concave/convex functions yield similar results to the linear one after tuning hyperparameters.} function with increasing values over epochs, ranging from 3 to 7. The model is trained with a learning rate of $1 \times 10^{-4}$ and a batch size of 32, utilizing the Adam optimizer \cite{kingma2014adam}. All experiments are conducted using PyTorch with NVIDIA A100 GPUs.
We evaluate our methods on four datasets: TGIF-QA \cite{jang2017tgif}, NExT-QA \cite{xiao2021next}, MSVD-QA \cite{xu2017video}, and MSRVTT-QA \cite{xu2017video}. Specifically, TGIF-QA consists of four sub-tasks: \textit{Count}, \textit{FrameQA}, \textit{Action}, and \textit{Transition}. 
% For instance, \textit{Count} is formulated as regression, \textit{FrameQA} involves open-ended tasks (multi-class classification), while \textit{Action} and \textit{Transition} are multi-choice tasks (binary classification for each option). NExT-QA is a multi-choice dataset encompassing description, chronology, and causality.
% MSVD-QA and MSRVTT-QA are open-ended VideoQA datasets. Regarding evaluation metrics, we employ mean square error (MSE) for \textit{Count} and accuracy for the other sub-tasks and datasets.

\subsection{Comparisons with Existing Methods}

In this section, we compare our model with the existing methods. 
% , including the state of the art, i.e., 
{As we discussed in Related Work, we follow HQGA \cite{xiao2022video}, IGV \cite{li2022invariant}, and ATP \cite{buch2022revisiting}, and compare only with the methods without large-scale video-text pretraining for fair comparisons.} \imp{We carefully selected the compared methods based on specific criteria: 1) the state of the art in recent years such as MASN \cite{seo2021attend}, IGV \cite{li2022invariant}, and HQGA \cite{xiao2022video}; 2) the methods that report results on the corresponding datasets in the original paper. In summary, we compare the best and result-available methods on each dataset.}

\subsubsection{Results on TGIF-QA}

% We first compare our methods with the existing ones on TGIF-QA. 
The results on TGIF-QA are shown in Table \ref{res}(left). Note that UCLQA$_F$/UCLQA$_P$ in the table represents our model trained with feature/predictive-uncertainty-aware CL. As we can see from the results, both UCLQA$_F$/UCLQA$_P$ achieve state-of-the-art performance on all sub-tasks except for \textit{FrameQA}. The improvements on the three sub-tasks are significant compared with the previous methods. As for \textit{FrameQA}, HQGA \cite{xiao2022video} achieves the best results for its hierarchical structure that is effective in capturing the global dependencies, and such global dependencies are crucial to \textit{FrameQA}. Note that the state-of-the-art method, IGV \cite{li2022invariant}, is not effective on \textit{FrameQA} and achieves the worst performance. 
% We assume the reason is that IGV struggles to pinpoint the causal scenes because the videos in TGIF-QA are relatively short.
Besides, UCLQA$_F$ and UCLQA$_P$ have similar performance on all sub-tasks of the TGIF-QA dataset.

\subsubsection{Results on NExT-QA}

We compare our methods with previous ones on NExT-QA. The results (accuracy on the validation/testing set) are shown in Table \ref{res}(right). As we can see from the results, our UCLQA$_F$ achieves better performance than the state-of-the-art methods, i.e., IGV \cite{li2022invariant}, ATP \cite{buch2022revisiting} and HQGA \cite{xiao2022video}. Specifically, the improvement in the accuracy on descriptive questions are significant compared to previous results. As for UCLQA$_F$, it further promotes the performance to a noticeable extent compared to HQGA and UCLQA$_F$. Besides, the improvement on causal questions is noteworthy. Nevertheless, our methods obtain inferior results on temporal questions. We assume the reason is that MASN models the relations among objects across frames, which diminishes the impact of temporal information on answer prediction\footnote{Please find the ablation study and more analysis in the supplementary material.}. 

% \subsubsection{Results on MSVD-QA and MSRVTT-QA}

% We have conducted a comparative analysis of our methodologies against previous approaches using the MSVD-QA \cite{xu2017video} and MSRVTT-QA \cite{xu2017video} datasets, with the results presented in Table \ref{msvd}. Notably, UCLQA$_P$ and IGV \cite{li2022invariant} jointly achieve the highest performance on MSVD-QA, whereas UCLQA$_F$ exhibits slightly lower results. Conversely, when considering MSRVTT-QA, both UCLQA$_P$ and UCLQA$_F$ trail behind IGV.
% This discrepancy may be attributed to the datasets' inherent properties. MSRVTT-QA features clear causal relations, which IGV capitalizes on by effectively leveraging invariant grounding to identify pivotal scenes for causal relation reasoning.
\begin{table}

\caption{(left) The results on TGIF-QA. The metric is MSE for \textit{Count} and is accuracy (\%) for others. $^*$ means the result is re-implementation with the official codes. UCLQA$_F$/UCLQA$_P$ represents our model trained with feature/predictive-uncertainty-aware CL. (right) The results on NExT-QA. The metric is accuracy (\%). MASN$^*$ is re-implementation.}
\label{res}
\vspace{1em}
\begin{tabular}{ccc}
\bmvaHangBox{
\resizebox{0.47\columnwidth}{!}{\begin{tabular}{@{}lllll@{}}
\toprule
                       Models            & \textit{Count}$\downarrow$ & \textit{FrameQA}$\uparrow$ & \textit{Action}$\uparrow$ & \textit{Trans.}$\uparrow$ \\ \midrule
B2A \cite{park2021bridge}          & 3.71                       & 57.5                       & 75.9                      & 82.6                      \\
HAIR \cite{liu2021hair}            & 3.88                       & 60.2                       & 77.8                      & 82.3                      \\
HOSTER \cite{dang2021hierarchical} & 4.13                       & 58.2                       & 75.6                      & 82.1                      \\
MASN \cite{seo2021attend}          & 3.64$^*$                       & 58.5$^*$                       & \underline{82.1}$^*$            & 85.7$^*$                  \\
HQGA \cite{xiao2022video}          & 3.97$^*$                   & \textbf{61.3}              & 76.9                      & 85.6                      \\
IGV \cite{li2022invariant}         & \underline{3.67}$^*$             & 52.8$^*$                   & 78.5$^*$                  & 85.7$^*$                  \\ \midrule
UCLQA$_F$                          & \textbf{3.22}              & \underline{60.5}                 & \textbf{84.0}             & \underline{87.7}                \\
UCLQA$_P$                          & \textbf{3.22}              & 60.4                       & \textbf{84.0}             & \textbf{87.8}             \\ \bottomrule
\end{tabular}}
}&
\bmvaHangBox{
\resizebox{0.47\columnwidth}{!}{\begin{tabular}{@{}lccccc@{}}
\toprule
\multirow{2}{*}{Models}         & \multirow{2}{*}{Val.} & \multicolumn{4}{c}{Testing}                                      \\ \cmidrule(l){3-6} 
                                &                      & Causal        & Temp.         & Descrip.      & All           \\ \midrule
CoMem \cite{gao2018motion}      & 44.2                 & 45.9          & 50.0          & 54.4          & 48.5          \\
HCRN \cite{le2020hierarchical}  & 48.2                 & 47.1          & 49.3          & 54.0          & 48.8          \\
HME \cite{fan2019heterogeneous} & 48.1                 & 46.8          & 48.9          & 57.4          & 49.2          \\
MASN$^*$ \cite{seo2021attend} &50.8&47.7&49.4&57.8&49.8\\
HGA \cite{jiang2020reasoning}   & 49.7                 & 48.1          & 49.1          & 57.8          & 50.0          \\
IGV \cite{li2022invariant}      & ---                  & 48.6          & \underline{51.7}    & 59.6          & 51.3          \\
ATP \cite{buch2022revisiting} &---&48.6&49.3&\textbf{65.0}&51.5\\
HQGA \cite{xiao2022video}       & 51.4                 & 49.0          & \textbf{52.3} & 59.4          & \underline{51.8}    \\ \midrule
UCLQA$_F$                       & \underline{51.6}           & \underline{49.4}    & 50.6          & \underline{61.9} & \underline{51.8}    \\
UCLQA$_P$                       & \textbf{52.3}        & \textbf{50.3} & 50.5          & 61.8    & \textbf{52.2} \\ \bottomrule
\end{tabular}}
}\\
\end{tabular}

\end{table}

\section{Conclusion}

In this paper, we propose a novel uncertainty-aware CL framework for VideoQA, where difficulty is gauged using uncertainty as a measure. To capture data uncertainty and mitigate its negative impact, we present a probabilistic modeling for VideoQA. Specifically, VideoQA is reformulated as a stochastic computation graph, wherein hidden representations of videos and questions become stochastic variables. Within this probabilistic modeling framework, we define feature uncertainty and predictive uncertainty to guide curriculum learning. In practice, we seamlessly integrate the VideoQA model into our framework and conduct experiments to demonstrate the superiority of our methods. The results illustrate that our approach achieves state-of-the-art performance across multiple datasets, while also providing meaningful uncertainty quantification for VideoQA.

\section{Acknowledgment}
This research was supported by the Australian Government through the Australian Research Council's DECRA funding scheme (Grant No.: DE250100030).

\newpage

\setcounter{section}{0}

\section*{Supplementary Material}

\section{Video-Question Interaction Module and Appearance-Motion Fusion Module}
\label{s2}

In this section, we elaborate on the video-question interaction module and appearance-motion fusion module in MASN \cite{seo2021attend}. Since the $K$ sets of visual features are processed identically, we omit the sampling index $k$ for simplicity.
After obtaining the visual representation $\mathcal{R}=\left\{\bm{r}^{z}\right\}_{z=1}^{Z}$ (for either appearance feature or motion feature) and question encoding $\left\{\bm{w}^{j}\right\}_{j=1}^{J}$, we combine each of the sets into a matrix, which is denoted as $\bm{R}\in\mathbb{R}^{Z\times d}$ and $\bm{W}\in\mathbb{R}^{J\times d}$. The video-question interaction is performed using the
bilinear attention network (BAN) \cite{kim2018bilinear} as follows,
\begin{equation}
\bm{B}_i=\bm{1}\cdot{\rm BAN}_i(\bm{B}_{i-1},\bm{R};\bm{C}_i)^{\mathrm{T}}+\bm{B}_{i-1},i=1,\cdots,4,
\end{equation}
where $\bm{B}_0=\bm{W}$, $\bm{1}=[1,1,\cdots,1]^{\mathrm{T}}\in\mathbb{R}^J$, and $\bm{C}_i$ is the attention map. Four blocks of BAN are applied, and the final result is the video-question interacted representation, which is denoted as $\bm{Q}\in\mathbb{R}^{J\times d}$.

The above stream is applied to appearance features and motion features in parallel. Then two types of  video-question representation, $\bm{Q}_a$ for appearance and $\bm{Q}_m$ for motion, are obtained.
$\bm{Q}_a$ and $\bm{Q}_m$ are fused with the motion-appearance-centered attention. Specifically, three types of attention are computed as follows,
\begin{align}
&\bm{P}_a = \mathrm{Attn}(\bm{U},\bm{Q}_a,\bm{Q}_a),\\
&\bm{P}_m = \mathrm{Attn}(\bm{U},\bm{Q}_m,\bm{Q}_m),\\
&\bm{P}_{mix} = \mathrm{Attn}(\bm{U},\bm{U},\bm{U}),
\end{align}
where $\bm{U}=[\bm{Q}_a,\bm{Q}_m]\in\mathbb{R}^{2J\times d}$, and $\mathrm{Attn}(\bm{X},\bm{Y},\bm{Z})$ is the scaled dot-product attention \cite{vaswani2017attention} with $\bm{X},\bm{Y},\bm{Z}$ as query, key, and value, respectively. Then, a residual connection \cite{he2016deep} and a LayerNorm \cite{ba2016layer} are further applied as follows,
\begin{equation}
\bm{Z}_x=\mathrm{LayerNorm}\left(\bm{P}_x+\bm{U}\right),
\end{equation}
where $x=a$, $m$, or $mix$, and $\bm{Z}_a/\bm{Z}_m/\bm{Z}_{mix}\in\mathbb{R}^{2J\times d}$ is the appearance-centered/motion-centered/mix attention. We then fused the three attention results with the guidance of questions as follows,
\begin{align}
&\bm{Z}=\sum_{x}\mathrm{softmax}_x\left(\frac{\bm{z}_x^{\mathrm{T}}\bm{w}^J}{\sqrt{d}}\right)\bm{Z}_x,\\
&\bm{O}=\mathrm{LayerNorm}\left(\bm{Z}+\mathrm{MLP}(\bm{Z})\right),
\end{align}
where $\bm{z}_x\in\mathbb{R}^{d}$ is the sum of $\bm{Z}_x$ along the first dimension. Finally, the video-question feature is computed by aggregating $\bm{O}\in\mathbb{R}^{2J\times d}$ along the first dimension as follows,
\begin{equation}
\bm{s}=\sum_{j}\mathrm{softmax}_j\left(\mathrm{MLP}(\bm{O}^j)\right)\bm{O}^j,
\end{equation}
where $\bm{O}^j\in\mathbb{R}^{d}$ is the $j$-th row of $\bm{O}$, $\mathrm{MLP}(\cdot)$ projects $\bm{O}^j$ to a scale. $\bm{s}\in\mathbb{R}^{d}$ is the fused feature for answer prediction.

\begin{algorithm}[tbp]
\caption{Uncertainty-aware Curriculum Learning for \textcolor{blue}{Classification}/\textcolor{red}{Regression}-based VideoQA}\label{alg}
\KwData{VideoQA training set $\left\lbrace (V_i,Q_i,a_i)\right\rbrace_{i=1}^D$.}
\KwData{Video encoder $F_{\phi}$, question encoder $G_{\psi}$, VQ interaction and answer decoder $H_{\theta}$.}
\KwData{Number of training epoch $E$, scheduler $\lambda(e)$, batch size $B$, learning rate $\gamma$, sampling times $K$.}
\For{$e=1..E$}{
\While{not done}{
Sample batch of data  $\left\lbrace (V_b,Q_b,a_b)\right\rbrace_{b=1}^B$\\
\For{$b=1..B$}{
$(\mu_m,\sigma_m)=F_{\phi}(V_b)$\\
$(\mu_n,\sigma_n)=G_{\psi}(Q_b)$\\
\For{$k=1..K$}{
Sample $\epsilon_m,\epsilon_n$ from $\mathcal{N}(0,1)$\\
$m_k=\mu_m+\sigma_m\epsilon_m$\\
$n_k=\mu_n+\sigma_n\epsilon_n$\\
$
\begin{cases}
\color{blue}\textnormal{\textit{cls}:}&  \color{blue}p^k=H_{\theta}(m_k,n_k)\\
\color{red}\textnormal{\textit{reg}:}&\color{red}\mu_k,\sigma^2_k=H_{\theta}(m_k,n_k)
\end{cases}
$\\
% $/\enspace\color{blue} p^k=H_{\theta}(m_k,n_k)$\\
% $\setminus \enspace  \color{red}\mu_k,\sigma^2_k=H_{\theta}(m_k,n_k)$\\
}

$
\begin{cases}
\color{blue}\textnormal{\textit{cls}:}& \color{blue}p^b=\frac{1}{K}\sum_{k}p^k \\
\color{red}\textnormal{\textit{reg}:}& \color{red}\mu_b=\frac{1}{K}\sum_{k}\mu_k
\end{cases}
$\\
% $/\enspace\color{blue}p^b=\frac{1}{K}\sum_{k=1}^Kp^k$\\
% $\setminus \enspace\color{red}\mu_b=\frac{1}{K}\sum_{k}\mu_k$\\

$
\begin{cases}
\color{blue}\textnormal{\textit{cls}:}& \color{blue}U_b=-\sum_{c}p^b_c\log p^b_c \\
\color{red}\textnormal{\textit{reg}:}& \color{red}U_b=\frac{1}{K}\sum_{k}(\mu_k-\mu_b)^2
\end{cases}
$\\

% $/\enspace\color{blue}U_b=-\sum_{c=1}^Cp^b_c\log p^b_c$\\
% $\setminus \enspace\color{red}U_b=\frac{1}{K}\sum_{k=1}^K(\mu_k-\mu_b)^2$\\
$w_b=\mathrm{Detach}\left(1-\sigma \left(\frac{\mathrm{BN}(U_b)}{\lambda (e)}\right)\right)$\\

$
\begin{cases}
\color{blue}\textnormal{\textit{cls}:}& \color{blue}l_b=-\frac{1}{K}\sum_{k}\log p^k_{a_b} \\
\color{red}\textnormal{\textit{reg}:}& \color{red}l_b=\frac{1}{K}\sum_{k}\left(\frac{(\mu_k-a_b)^2}{\sigma_k^2}+\log\sigma_k^2\right)
\end{cases}
$\\

% $/\enspace\color{blue}l_b=-\sum_{k}\log p^k_{a_b}$\\
% $\setminus \enspace\color{red}l_b=\sum_{k}\left(\frac{1}{\sigma_k^2}(\mu_k-a_b)^2+\log\sigma_k^2\right)$
}
$\mathcal{L}=\frac{1}{B}\sum_{b} w_bl_b$\\
$(\phi,\psi,\theta)=(\phi,\psi,\theta)-\gamma\nabla_{(\phi,\psi,\theta)}\mathcal{L}$\\
}
}
\end{algorithm}

\section{Pseudo Code of Uncertainty-Aware Curriculum Learning for VideoQA}
\label{s1}

Alg. \ref{alg} shows the pseudo code of our uncertainty-aware curriculum learning for VideoQA. Note that both the classification-based (CB) VideoQA and regression-based (RB) VideoQA are illustrated in the algorithm: keep the \textcolor{blue}{blue} lines (the tops of Line 11, 13, 14, 16) and discard the \textcolor{red}{red} ones (the bottoms of Line 11, 13, 14, 16) for CB VideoQA, while do the reverse for RB VideoQA. Besides, we use predictive-uncertainty for CB VideoQA and feature-uncertainty for RB VideoQA for examples, where the uncertainty can be easily replaced with the other type. The KL divergence regularization for the feature distributions  in the loss is omitted for simplicity. 

\section{Results on MSVD-QA and MSRVTT-QA}
\label{s3}

We also compare our methods with previous ones on MSVD-QA \cite{xu2017video} and MSRVTT-QA \cite{xu2017video}. The results are shown in Table \ref{msvd}. Specifically, our UCLQA$_P$ and IGV \cite{li2022invariant} achieve the best performance on MSVD-QA, while our UCLQA$_F$ achieves slightly lower result. As for MSRVTT-QA, both our UCLQA$_P$ and UCLQA$_F$ are inferior to IGV. We assume the reason is that the causal relations in this dataset are clear, and such a property is fully exported by IGV where the invariant grounding can accurately find the crucial scenes for casual relation reasoning. 

\begin{table}[tbp]
\caption{The accuracy (\%) on MSVD-QA and  MSRTT-QA.} 
\label{msvd}
\center
\begin{tabular}{@{}lcc@{}}
\toprule
Method                             & MSVD-QA       & MSRTT-QA      \\ \midrule
B2A \cite{park2021bridge}          & 37.2          & 36.9          \\
HAIR \cite{liu2021hair}            & 37.5          & 36.9          \\
MASN \cite{seo2021attend}          & 38.0          & 35.3          \\
DualVGR \cite{wang2021dualvgr}     & 39.0          & 35.5          \\
HOSTER \cite{dang2021hierarchical} & 39.4          & 35.9          \\
IGV \cite{li2022invariant}         & \textbf{40.8} & \textbf{38.3} \\ \midrule
UCLQA$_F$                          & {\ul 40.6}    & {\ul 37.3}    \\
UCLQA$_P$                          & \textbf{40.8} & {\ul 37.3}    \\ \bottomrule
\end{tabular}
\end{table}

\begin{table}[tbp]
\caption{The results of the uncertainty-aware CL equipped with  different types of uncertainty quantification. The improvements are highlighted in {\color[HTML]{036400}green}.} 
\label{unc}
\center
\begin{tabular}{cll}
\toprule
Uncertainty & \multicolumn{1}{c}{\textit{Count}$\downarrow$} & \multicolumn{1}{c}{\textit{FrameQA}$\uparrow$} \\ \hline
\rowcolor[HTML]{C0C0C0} Baseline    &  3.64 & 58.5             \\
$U_{T}$           & \underline{3.50}   {\color[HTML]{036400}(-1.4)}   &   59.7   {\color[HTML]{036400}(+1.2)}   \\
$U_{F}$           &      \textbf{3.22} {\color[HTML]{036400}(-4.2)}& \textbf{60.5}  {\color[HTML]{036400}(+2.0)}      \\
$U_{P}$            &      \textbf{3.22} {\color[HTML]{036400}(-4.2)}&\underline{60.4}  {\color[HTML]{036400}(+1.9)}      \\ \bottomrule
\end{tabular}
\end{table}

\section{More Analysis}
\label{ma}

% \subsubsection{Probabilistic Modeling and Uncertainty-Aware Curriculum Learning}

\subsection{Ablation Study} 

We conduct ablation studies to show the impact of the proposed probabilistic modeling and uncertainty-aware CL. {The baseline model is constructed by discarding probabilistic modeling, where the visual representations (including appearance and motion) are deterministic variables instead of random variables, and CL is not applied in the training process.} Besides, we also evaluate the model with probabilistic modeling but trained without CL. The results on TGIF-QA are shown in Table \ref{ab1}. As we can from the results, the baseline model achieves considerable results. Furthermore, the probabilistic modeling improves the performance on all sub-tasks of TGIF-QA, which means modeling the uncertainty in the visual feature space is beneficial to capturing robust spatial-temporal dependencies in videos for question answering. 
% Based on the probabilistic modeling, we compute two types of uncertainty (feature uncertainty and predictive uncertainty) and use each of them for CL. 
Furthermore, feature-uncertainty-aware CL (CL$_F$) and predictive-uncertainty-aware CL (CL$_P$) both improve the performance on all sub-tasks. Specifically, the improvements on \textit{Count}, \textit{FrameQA}, and \textit{Action} are significant. We assume the reason for the less significant improvement on \textit{Transition} is that the variance of the uncertainty distribution of this dataset is small, which diminishes the impact of CL.

\begin{table*}[tbp]
\caption{The results of the model trained by different CL methods. 
% \textit{C}/\textit{F}/\textit{A}/\textit{T} represents the subset of TGIF-QA: \textit{Count} / \textit{FrameQA} / \textit{Action} / \textit{Transition}. 
CL$_H$ is the original SPL \cite{kumar2010self}. CL$_L$ is the SPL with linear regularizer \cite{jiang2014easy}. CL$_F$/CL$_P$ represents feature/predictive-uncertainty-aware CL. The improvements/declines in performance are highlighted in {\color[HTML]{036400}green}/{\color[HTML]{CB0000}red}.} 
\label{abcl}
\center
\resizebox{\columnwidth}{!}{\begin{tabular}{cllllll}
\toprule
\multirow{2}{*}{CL} &
  \multicolumn{4}{c}{TGIF-QA} &
  \multicolumn{1}{c}{\multirow{2}{*}{MSVD$\uparrow$}} &
  \multicolumn{1}{c}{\multirow{2}{*}{MSRVTT$\uparrow$}} \\ \cmidrule(lr){2-5}
 &
  \multicolumn{1}{c}{\textit{Count}$\downarrow$} &
  \multicolumn{1}{c}{\textit{FrameQA}$\uparrow$} &
  \multicolumn{1}{c}{\textit{Action}$\uparrow$} &
  \multicolumn{1}{c}{\textit{Trans.}$\uparrow$} &
   &
   \\ \hline
\rowcolor[HTML]{C0C0C0}  / & \underline{3.33}    & 59.7          & 82.8          & 87.2          & 38.9          & 36.1          \\
CL$_H$   & 3.59  {\color[HTML]{CB0000}(+0.26)}        & 59.9 {\color[HTML]{036400}(+0.2)}         & \underline{83.0}  {\color[HTML]{036400}(+0.2)}  & 86.6   {\color[HTML]{CB0000}(-0.6)}       & 38.8   {\color[HTML]{CB0000}(-0.1)}       & \underline{36.4}  {\color[HTML]{036400}(+0.3)}  \\
CL$_L$   &3.46 {\color[HTML]{CB0000}(+0.13)}&60.0 {\color[HTML]{036400}(+0.3)}&82.8 (+0.0)&87.1 {\color[HTML]{CB0000}(-0.1)}&39.4 {\color[HTML]{036400}(+0.5)}& \underline{36.4} {\color[HTML]{036400}(+0.3)}\\\midrule
CL$_F$   & \textbf{3.22}  {\color[HTML]{036400}(-0.11)}& \textbf{60.5}  {\color[HTML]{036400}(+0.8)}& \textbf{84.0}  {\color[HTML]{036400}(+1.2)}& \underline{87.7}   {\color[HTML]{036400}(+0.5)}  & \underline{40.6}    {\color[HTML]{036400}(+1.7)} & \textbf{37.3}  {\color[HTML]{036400}(+1.2)}\\
CL$_P$   & \textbf{3.22}  {\color[HTML]{036400}(-0.11)}& \underline{60.4}     {\color[HTML]{036400}(+0.7)}& \textbf{84.0}  {\color[HTML]{036400}(+1.2)}& \textbf{87.8}  {\color[HTML]{036400}(+0.6)}& \textbf{40.8}  {\color[HTML]{036400}(+1.9)}& \textbf{37.3}  {\color[HTML]{036400}(+1.2)}\\ \bottomrule
\end{tabular}}
\end{table*}

\begin{table}[tbp]
\setlength{\tabcolsep}{5pt}
\caption{The results of ablation studies on TGIF-QA. PM is the abbreviation for probabilistic modeling. CL$_F$/CL$_P$ represents feature/predictive-uncertainty-aware curriculum learning. The improvements are highlighted in {\color[HTML]{036400}green}.} 
\label{ab1}
\center
\begin{tabular}{cccllll}
\toprule
PM & CL$_F$ & CL$_P$ & \multicolumn{1}{c}{\textit{Count}$\downarrow$} & \multicolumn{1}{c}{\textit{FrameQA}$\uparrow$} & \multicolumn{1}{c}{\textit{Action}$\uparrow$} & \multicolumn{1}{c}{\textit{Trans.}$\uparrow$} \\ \hline
\rowcolor[HTML]{C0C0C0}  &  &  & 3.64 & 58.5 & 82.1 & 85.7 \\
\checkmark &  &  & 3.33 {\color[HTML]{036400}(-0.31)} & 59.7 {\color[HTML]{036400}(+1.2)}& \underline{82.8} {\color[HTML]{036400}(+0.7)}& 87.2 {\color[HTML]{036400}(+1.5)} \\
\checkmark & \checkmark &  & \textbf{3.22} {\color[HTML]{036400}(-0.42)} & \textbf{60.5} {\color[HTML]{036400}(+2.0)}& \textbf{84.0} {\color[HTML]{036400}(+1.9)}& \underline{87.7} {\color[HTML]{036400}(+2.0)}\\
\checkmark &  & \checkmark & \textbf{3.22} {\color[HTML]{036400}(-0.42)} & \underline{60.4} {\color[HTML]{036400}(+1.9)}& \textbf{84.0} {\color[HTML]{036400}(+1.9)}& \textbf{87.8} {\color[HTML]{036400}(+2.1)}\\ \bottomrule
\end{tabular}
\end{table}

\subsection{Curriculum Learning} 

We compare our uncertainty-aware curriculum learning with the original SPL (CL$_H$) \cite{kumar2010self} and the SPL with linear regularizer (CL$_L$) \cite{jiang2014easy}. The results on three datasets are shown in Table \ref{abcl}. \textit{The baseline model is our model with only probabilistic modeling (the second row in Table \ref{ab1}).} For fair comparisons, we also apply probabilistic modeling to the model trained with CL$_H$ and CL$_L$. As we can see from the results, CL$_H$ makes little difference to the performance on MSVD-QA, and it has a negative impact on \textit{Count} and \textit{Transition}. Besides, the improvements on \textit{FrameQA}, \textit{Action}, and MSRVTT-QA are marginal. As for CL$_L$, it brings obvious improvement only on MSVD-QA, while the impact on other datasets is either little or negative. In contrast, the proposed CL$_P$ and CL$_F$ both achieve obvious improvements on all sub-tasks and datasets, which shows the superiority of our methods over traditional SPL.

\subsection{Uncertainty Quantification}
\label{s4}

We have conducted a comparison of the uncertainty-aware curriculum learning, employing distinct forms of uncertainty quantification. Specifically, we have juxtaposed the uncertainty quantification technique proposed in \cite{kendall2017uncertainties} ($U_T$) with our two variations of uncertainty ($U_F$ and $U_P$) for both regression and classification tasks on \textit{Count} and \textit{FrameQA}, respectively. The results, as depicted in Table \ref{unc}, are contrasted against a baseline model trained without curriculum learning.
From the outcomes, it is evident that curriculum learning enhanced with $U_T$ yields improvements in performance for both the regression task (\textit{Count}) and the classification task (\textit{FrameQA}). This underscores the efficacy of our uncertainty-aware curriculum learning framework, demonstrating its adaptability to different uncertainty quantification methodologies. Furthermore, our proposed types of uncertainty provide even more significant performance gains compared to $U_T$, largely attributed to the added benefit of probabilistic modeling.

\begin{table}[tbp]
\caption{The results of the uncertainty-aware CL equipped with  different types of uncertainty quantification. The improvements are highlighted in {\color[HTML]{036400}green}.} 
\label{unc}
\center
\begin{tabular}{cll}
\toprule
Uncertainty & \multicolumn{1}{c}{\textit{Count}$\downarrow$} & \multicolumn{1}{c}{\textit{FrameQA}$\uparrow$} \\ \hline
\rowcolor[HTML]{C0C0C0} Baseline    &  3.64 & 58.5             \\
$U_{T}$           & \underline{3.50}   {\color[HTML]{036400}(-1.4)}   &   59.7   {\color[HTML]{036400}(+1.2)}   \\
$U_{F}$           &      \textbf{3.22} {\color[HTML]{036400}(-4.2)}& \textbf{60.5}  {\color[HTML]{036400}(+2.0)}      \\
$U_{P}$            &      \textbf{3.22} {\color[HTML]{036400}(-4.2)}&\underline{60.4}  {\color[HTML]{036400}(+1.9)}      \\ \bottomrule
\end{tabular}
\end{table}

\begin{table}[tbp]
\caption{\imp{The comparisons (accuracy \%) between the models trained without/with our UCL framework. $^*$IGV is trained by simple cross-entropy loss for fair comparisons.} }
\label{gen}
\center
\begin{tabular}{@{}lcccc@{}}
\toprule
\multirow{2}{*}{Models} & \multicolumn{2}{c}{TGIF-\textit{Action}}             & \multicolumn{2}{c}{MSVD-QA}                          \\ \cmidrule(l){2-3} \cmidrule(l){4-5} 
                        & w/o UCL           & w/ UCL                           & w/o UCL           & w/ UCL                           \\ \midrule
HGA                     & 76.0 & 78.5 {\color[HTML]{036400}(+2.5)} & 33.1 & 34.7 {\color[HTML]{036400}(+1.6)} \\
HQGA                    & 76.9 & 78.8 {\color[HTML]{036400}(+1.9)} & 39.7 & 41.5 {\color[HTML]{036400}(+1.8)} \\
IGV$^*$                 & 78.5 & 79.6 {\color[HTML]{036400}(+1.1)} & 35.6& 37.0 {\color[HTML]{036400}(+1.4)} \\
MASN                    & 82.1 & 84.0 {\color[HTML]{036400}(+1.9)} & 38.0 & 40.8 {\color[HTML]{036400}(+2.8)} \\ \bottomrule
\end{tabular}
\end{table}

\begin{figure}[tp]
\centering
\includegraphics[width=0.7\columnwidth]{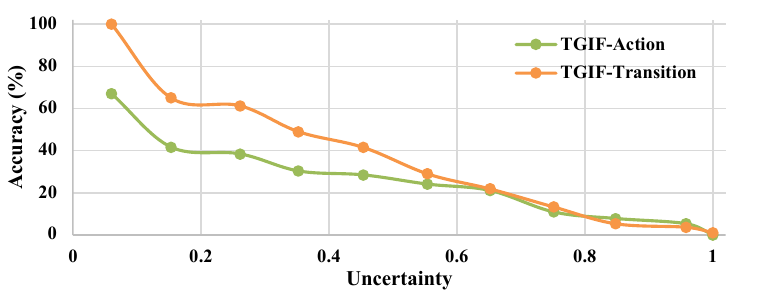}
\caption{\imp{The Uncertainty-Accuracy curves for TGIF-\textit{Action} and TGIF-\textit{Transition}.}}
\label{con-acc}
\end{figure}

\subsection{Generalization Ability} 

\imp{Demonstrating the robustness of our approach in terms of generalization across various VideoQA models, we also applied it to HGA, HQGA, and IGV. The outcomes, presented in Table \ref{gen}, consistently showcase substantial enhancements across these models, both for TGIF-\textit{Action} (multi-choice) and MSVD-QA (open-ended) datasets. Notably, a straightforward modification results in accuracy improvements exceeding 1.5\% in the majority of cases. These findings underscore the versatility of our UCL framework, showcasing its applicability across a diverse range of VideoQA models to achieve superior performance.}

\subsection{Quality of Uncertainty} 
\imp{To quantitatively analyze the accuracy of our uncertainty, we discretize the predictive uncertainty (normalized to [0,1]) into ten levels and compute the accuracy within each level. The Uncertainty-Accuracy curve on TGIF-\textit{Action} and TGIF-\textit{Transition} is shown in Figure \ref{con-acc}. We have observed a negative correlation between confidence and accuracy, which supports the validity of our uncertainty estimation. Therefore, a promising application of our model is that it can quantify the uncertainty in VideoQA data and assess the difficulty of videos and QA pairs, which can greatly accelerate the collection of challenging data. }

%  \begin{figure*}[tbp]
%      \centering
%      \begin{subfigure}[b]{\textwidth}
%          \centering
%          \includegraphics[width=\textwidth]{pred.pdf}
%          \caption{ Examples of high predictive uncertainty. The uncertainty of each option is also provided (normalized to $[0,1]$).}
%          \label{pred}
%      \end{subfigure}
     
%      \begin{subfigure}[b]{\textwidth}
%          \centering
%          \includegraphics[width=\textwidth]{feat.pdf}
%          \caption{Examples of high feature uncertainty.}
%          \label{feat}
%      \end{subfigure}
%         \caption{Examples of high uncertainty. The predictions are in \colorbox[RGB]{251,229,214}{orange}, while the correct answers are in \colorbox[RGB]{226,240,217}{green}.}
%         \label{vis}
% \end{figure*}

\begin{figure}
\begin{tabular}{cc}
\bmvaHangBox{\includegraphics[width=0.98\textwidth]{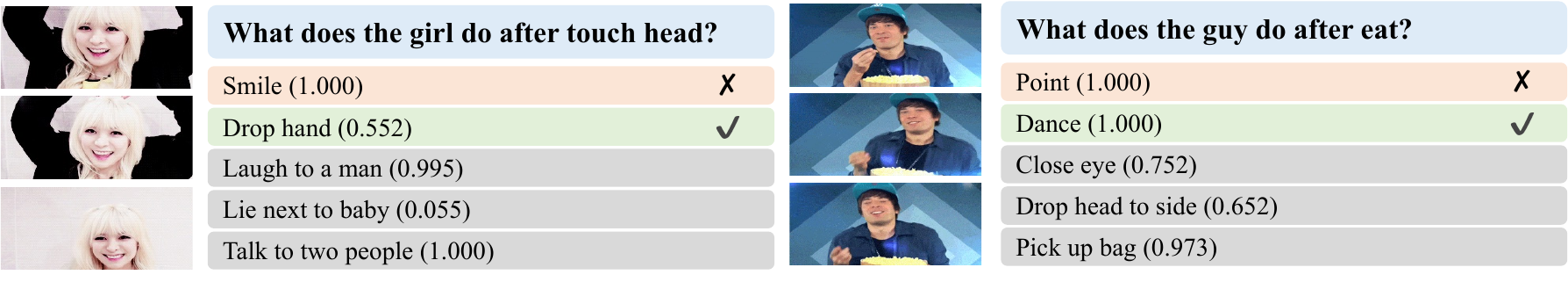}\\[-0.1pt]\\
\bmvaHangBox{\includegraphics[width=0.98\textwidth]{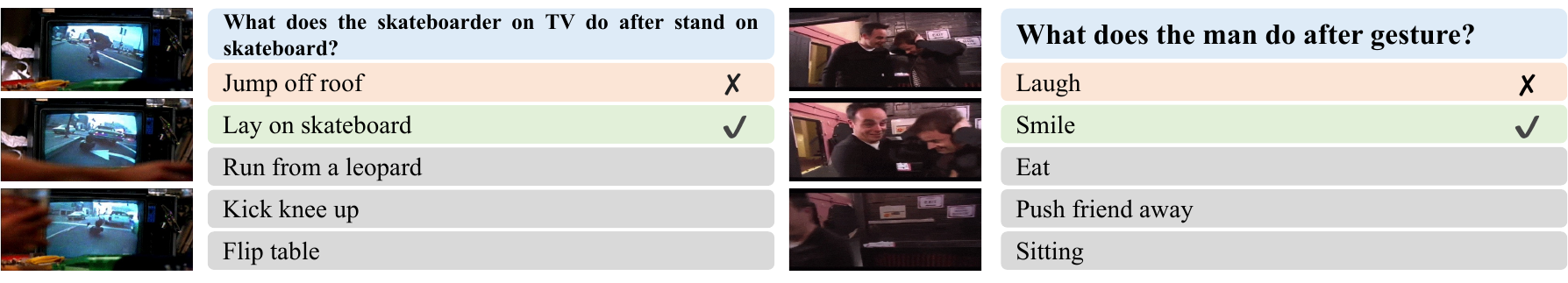}}}\\
\end{tabular}
\vspace{0.5em}
\caption{(Up) Examples of high predictive uncertainty. The uncertainty of each option is also provided (normalized to $[0,1]$). (Down) Examples of high uncertainty. The predictions are in \colorbox[RGB]{251,229,214}{orange}, while the correct answers are in \colorbox[RGB]{226,240,217}{green}.}
\label{vis}
\end{figure}

\subsection{Uncertainty Visualization} 
Figure \ref{vis}(Up) shows some examples of high predictive uncertainty in TGIF-\textit{Transition}, where the uncertainty of each option is provided. As the figure shows, the wrong predictions are of high predictive uncertainty, which means our model has less confidence in its predictions and thus makes incorrect choices. Furthermore, there exists obvious ambiguity in the high-uncertainty videos, e.g., in the right video of Figure \ref{vis}(Up), while the man is dancing, he also performs a pointing action that is very obvious, so it makes sense that our model has low confidence in both ``Dance" and ``Point".
% and struggles to choose from the ambiguous answers. 
Figure \ref{vis}(Down) illustrates some examples of high feature uncertainty in TGIF-\textit{Transition}. As we can see from the figure, the videos of high feature uncertainty are generally in poor visual quality. For example, the question of the left video in Figure \ref{vis}(Down) asks about the actions on TV, which are not clear enough to tell what actually happens. Besides, for the right video, the man moves very quickly, and it is hard to tell whether he is laughing or smiling. These examples demonstrate that poor visual quality results in high feature uncertainty, which could have a negative impact on the accuracy of the predictions.

\subsection{Hyper-parameter Analysis}
\label{s5}

\subsubsection{KL Divergence Weight $\alpha$} 

In our probabilistic modeling, we apply KL divergence regularization to the stochastic representations to prevent their degradation into deterministic ones, with the regularization strength balanced by a weight denoted as $\alpha$. The impact of varying $\alpha$ on the model trained with predictive-uncertainty-aware CL for NExT-QA is presented in Table \ref{next1}. The reported results include both validation and testing accuracy. As observed from the table, performance peaks at $\alpha=1\times 10^{-4}$, while larger values of $\alpha$ lead to decreased accuracy. This pattern is likely attributed to the high-quality videos in the NExT-QA dataset, where minor uncertainty exists in the feature space. Introducing strong regularization to the features might inadvertently compromise information integrity.
Similar analyses were conducted on TGIF-QA, MSVD, and MSRVTT datasets, yielding optimal results around $\alpha=1\times 10^{-1}$ or $\alpha=1\times 10^{-2}$ across various datasets or sub-tasks within TGIF-QA. This trend suggests that larger $\alpha$ values are advantageous for datasets with visually unsatisfactory videos (such as TGIF-QA), facilitating the learning of more robust representations.

\begin{table}[tbp]
\caption{The accuracy (\%) of different $\alpha$ on NExT-QA.} 
\label{next1}
\center
\begin{tabular}{@{}cccccc@{}}
\toprule
\multirow{2}{*}{$\alpha$} & \multirow{2}{*}{Val.} & \multicolumn{4}{c}{Testing}                                   \\ \cmidrule(l){3-6} 
                          &                       & Causal        & Temp.         & Descrip.      & All           \\ \midrule
$1\times10^{-5}$          & \underline{52.2}            & 48.4          & 49.9          & \textbf{62.6} & 51.2          \\
$5\times10^{-5}$          & \underline{52.2}            & \underline{49.7}    & \textbf{50.7} & 61.7          & 52.0          \\
$1\times10^{-4}$          & \textbf{52.3}         & \textbf{50.3} & \underline{50.5}    & 61.8          & \textbf{52.2} \\
$5\times10^{-4}$          & 51.8                  & \underline{49.7}    & 49.4          & \underline{61.9}    & 51.6          \\
$1\times10^{-3}$          & 51.1                  & 48.8          & 49.5          & 60.9          & 51.0          \\ \bottomrule
\end{tabular}
\end{table}

\begin{table}[tbp]
\caption{The results of different training scheduler on the \textit{Count} task of TGIF-QA. The MSE of validation/testing is presented.} 
\label{cnt}
\center
\begin{tabular}{@{}cccc@{}}
\toprule
 $S_1$\textbackslash $S_2$ & 5 & 6 & 7 \\ \midrule
1 & \textbf{3.24} / \textbf{3.22}  & 3.33 / 3.35  & 3.26 / 3.30  \\
2 &  3.29 / 3.30 & 3.28 / 3.30  & 3.25 / 3.29  \\
3 &  3.26 / 3.25 & 3.31 / 3.30 & 3.32 / 3.33  \\ \bottomrule
\end{tabular}
\end{table}

\subsubsection{Training Scheduler $\lambda(e)$} 

The training scheduler $\lambda(e)$ plays a pivotal role in controlling the rate at which the difficulty level changes in the curriculum learning process. In this study, a linear increasing function with respect to the epoch is adopted, given by:
\begin{equation}
\lambda(e) = \frac{S_2-S_1}{E-1}e+S_1,
\end{equation}
where $E$ corresponds to the total number of training epochs ($e=0,1,\cdots,E-1$), and $S_1$ and $S_2$ ($S_1<S_2$) serve as hyper-parameters that govern the rate of difficulty adaptation. The results for different settings of $S_1$ (1, 2, 3) and $S_2$ (5, 6, 7) on the \textit{Count} task within TGIF-QA are displayed in Table \ref{cnt}, featuring the reported mean squared error (MSE) on the validation set. The outcomes reveal that the optimal performance is achieved when $S_1=1$ and $S_2=5$.
In our experimental procedures, the selection of the optimal $S_1$ and $S_2$ values was based on validation performance across all datasets (or sub-tasks within TGIF-QA). Generally, for smaller-scale datasets (or sub-tasks) like \textit{Count} and \textit{FrameQA}, smaller and more rapidly increasing scheduler values (e.g., $1\rightarrow5$) tend to yield better results. Conversely, for larger-scale datasets such as \textit{Transition} and NExT-QA, a larger, slower-increasing scheduler (e.g., $3\rightarrow7$) is preferred. This distinction arises from the observation that larger-scale data necessitate a longer duration to adapt to the evolving difficulty level effectively.

\begin{table}[tbp]
\caption{The results of models with different sampling times. \textit{C}/\textit{F}/\textit{A}/\textit{T} represents the subset of TGIF-QA: \textit{Count} / \textit{FrameQA} / \textit{Action} / \textit{Transition}.
UCLQA$_K$ represents the model with $K$ sampling times. The numbers of parameters are also reported.} 
\label{time}
\center
\begin{tabular}{@{}lllllc@{}}
\toprule
Model                      & \multicolumn{1}{c}{\textit{C}$\downarrow$} & \multicolumn{1}{c}{\textit{F}$\uparrow$} & \multicolumn{1}{c}{\textit{A}$\uparrow$} & \multicolumn{1}{c}{\textit{T}$\uparrow$} & \#Param. (M)\\ \midrule
MASN \cite{seo2021attend}  & 3.64$^*$                       & 58.5$^*$                       & \underline{82.1}$^*$            & 85.7$^*$                    &  \underline{25.7}  \\
HQGA \cite{xiao2022video}  & 3.97$^*$                   & \textbf{61.3}              & 76.9                      & 85.6                       &\textbf{11.0}         \\
IGV \cite{li2022invariant} & 3.67$^*$                   & 52.8$^*$                   & 78.5$^*$                  & 85.7$^*$                &   34.5      \\ \midrule
UCLQA$_1$                  & \underline{3.22}                 & \underline{60.4}                 & 83.4                      & 87.5                     & \multirow{4}{*}{27.3}    \\
UCLQA$_3$                  & 3.24                       & \underline{60.4}                 & 83.6                      & 87.6                          \\
UCLQA$_7$                  & \textbf{3.21}              & \underline{60.4}                 & \underline{83.9}                & \underline{87.6}                          \\
UCLQA$_{10}$               & \underline{3.22}                 & \underline{60.4}                 & \textbf{84.0}             & \textbf{87.8}                       \\ \bottomrule
\end{tabular}
\end{table}

\subsubsection{Sampling Times} 

Table \ref{time} presents the outcomes of our model with varied sampling times during inference on TGIF-QA (trained with predictive-uncertainty-aware CL). Throughout training, a consistent sampling time of 5 is employed across all sub-tasks. The results indicate that, during the testing phase, different sampling times have minimal impact on predictions across all sub-tasks except for \textit{Action}. This observation likely stems from the enhanced robustness of our model due to probabilistic modeling. Consequently, minor disruptions in the encoded visual representations exert limited influence on video-question interaction and answer prediction.
For  \textit{Action}, performance improves with more sampling times. However, the enhancement becomes marginal when the sampling time is large. Table \ref{time} also provides insights into model parameters. Notably, HQGA features the fewest parameters, while IGV boasts the most. In comparison, our model and MASN exhibit similar parameter counts. \textbf{Importantly, the sampling is performed after video encoding, and the subsequent repeated computation can be implemented in a parallel way instead of the serial one. Consequently, the inference speed remains relatively unaffected.} For instance, the inference time (in seconds) of UCLQA$_1$(MASN)/UCLQA$_3$/UCLQA$_7$/UCLQA$_{10}$ on \textit{Count} is 201/208/215/221 on an NVIDIA A100 GPU.

\section{Limitations}

Existing VideoQA methods exploit deep neural networks for video encoding and text encoding, where the encoded features are deterministic \cite{seo2021attend,xiao2022video,li2022invariant}. However, there exists inherent uncertainty in data such as noise, blur, and occlusion in videos. To address this issue, we propose probabilistic modeling where the representations are random variables, aiming to reduce the impact of uncertainty and measure the uncertainty. On the other hand, the model with feature-level probabilistic modeling might be less sensitive to the subtle changes in the videos. That is to say, the model trained by our framework might be less capable of distinguishing visually similar concepts, which could have a negative impact on fine-grained video understanding. We will focus on this issue in the future work. 

\begin{figure*}[t]
\begin{align}
&\log p_{\theta,\phi,\psi}(y|V,Q)\notag\\
&=\log \int_{m,n} p_{\theta,\phi,\psi}(y,m,n|V,Q)\mathrm{d}m\mathrm{d}n \notag\\
&=\log \int_{m,n} p_{\theta,\phi,\psi}(y,{m,n}|V,Q)\frac{q_{\phi}({m,n}|V,Q)}{q_{\phi}({m,n}|V,Q)}\mathrm{d}m\mathrm{d}n\notag\\
&=\log \mathbb{E}_{(m,n)\sim q_{\phi,\psi}({m,n}|V,Q)}\left[\frac{p_{\theta,\phi,\psi}(y,m,n|V,Q)}{q_{\phi,\psi}(m,n|V,Q)}\right] \notag\\
 &\geq \mathbb{E}_{(m,n)\sim q_{\phi,\psi}({m,n}|V,Q)}\left[\log\frac{p_{\theta,\phi,\psi}(y,m,n|V,Q)}{q_{\phi,\psi}(m,n|V,Q)}\right] \textnormal{(Jensen's inequality)}\notag\\
&= \mathbb{E}_{(m,n)\sim q_{\phi,\psi}({m,n}|V,Q)}\left[\log\frac{p_{\theta}(y|m,n)p(m,n)}{q_{\phi,\psi}(m,n|V,Q)p(V,Q)}\right] \notag\\
&=\mathbb{E}_{(m,n)\sim q_{\phi,\psi}({m,n}|V,Q)}\left[\log p_{\theta}(y|m,n)-\log\frac{q_{\phi,\psi}(m,n|V,Q)}{q(m,n)}-\log p(V,Q)\right] \notag\\
 &= \mathbb{E}_{(m,n)\sim q_{\phi,\psi}({m,n}|V,Q)}\left[\log p_{\theta}(y|m,n)\right]-\mathbb{E}_{(m,n)\sim q_{\phi,\psi}(m,n|V,Q)}\left[\log\frac{q_{\phi,\psi}(m,n|V,Q)}{q(m,n)}\right]\notag\\
 &-\mathbb{E}_{z\sim q_{\phi,\psi}(m,n|V,Q)}\left[\log p(V,Q)\right] \notag\\
&= \mathbb{E}_{(m,n)\sim q_{\phi,\psi}({m,n}|V,Q)}\left[\log p_{\theta}(y|m,n)\right]-D_{\mathrm{KL}}\left(q_{\phi,\psi}(m,n|V,Q)||p(m,n)\right)-\log p(V,Q)\notag\\
&= \mathbb{E}_{(m,n)\sim q_{\phi,\psi}({m,n}|V,Q)}\left[\log p_{\theta}(y|m,n)\right]-D_{\mathrm{KL}}\left(q_{\phi}(m|V)q_{\psi}(n|Q)||p(m)p(n)\right)-\log p(V)p(Q)\notag\\
&= \mathbb{E}_{(m,n)\sim q_{\phi,\psi}({m,n}|V,Q)}\left[\log p_{\theta}(y|m,n)\right]-D_{\mathrm{KL}}\left(q_{\phi}(m|V)||p(m)\right)\notag\\
&-D_{\mathrm{KL}}\left(q_{\psi}(n|Q)||p(n)\right)-\log p(V)p(Q) \label{elbo}
\end{align}
\end{figure*}

% \section{Visualization of Feature Uncertainty}
% \label{s6}

% Figure \ref{feat_eg} illustrates some examples of high feature uncertainty in TGIF-QA. As we can see from the figure, the videos of high feature uncertainty are generally in poor visual quality. For example, the question of the first video in Figure \ref{feat_eg} quires the actions on TV, which are not clear enough to tell what actually happens. Besides, for the video at the bottom-left, the man moves very quickly, and it is hard to tell whether he is laughing or smiling. These examples demonstrate that poor visual quality results in high feature uncertainty, which could have a negative impact on the accuracy of the predictions for VideoQA.

\section*{Appendix: Derivation of ELBO}

The detailed derivation of the evidence lower bound (ELBO) is shown in Eq. \ref{elbo}.

% \section{Future Work}

% Current VideoQA methodologies primarily leverage deep neural networks for deterministic video and text encoding, yielding encoded features \cite{seo2021attend,xiao2022video,li2022invariant}. However, inherent uncertainty is present in data due to factors like noise, blur, and occlusion within videos. To mitigate this challenge, we propose probabilistic modeling, where representations are treated as random variables. This approach aims to diminish the influence of uncertainty and quantify it effectively.
% Conversely, a model employing feature-level probabilistic modeling might exhibit reduced sensitivity to subtle changes in videos. In other words, a model trained within our framework might possess a diminished ability to differentiate visually similar concepts, potentially impacting fine-grained video comprehension. This specific concern will be a focal point in our forthcoming work.

\bibliography{egbib}
\end{document}